\pdfoutput=1
\documentclass[11pt]{article}
\usepackage[]{acl}
\usepackage{times}
\usepackage{latexsym}
\usepackage[T1]{fontenc}
\usepackage[utf8]{inputenc}
\usepackage{microtype}
\usepackage{inconsolata}
\usepackage{multirow}

\title{Mothman at SemEval-2024 Task 9:\\
An Iterative System for Chain-of-Thought Prompt Optimization}
\author{Alvin Po-Chun Chen \and Ray Groshan \and Sean von Bayern\\
  University of Colorado Boulder\\
  \texttt{\{alvin.chen, ray.groshan, sean.vonbayern\}@colorado.edu}}

\begin{document}
\maketitle
\begin{abstract}
Extensive research exists on the performance of large language models on logic-based tasks, whereas relatively little has been done on their ability to generate creative solutions on lateral thinking tasks. The \textsc{BrainTeaser} shared task tests lateral thinking and uses adversarial datasets to prevent memorization, resulting in poor performance for out-of-the-box models. We propose a system for iterative, chain-of-thought prompt engineering which optimizes prompts using human evaluation. Using this shared task, we demonstrate our system's ability to significantly improve model performance by optimizing prompts and evaluate the input dataset.\footnote{Our code can be found at \href{https://github.com/alvin-pc-chen/semeval_brainteaser}{https://github.com/alvin-pc-chen/semeval\_brainteaser}.}
\end{abstract}

\section{Introduction}
The ability for language models to reason or possess common sense knowledge has become a controversial topic with far-reaching implications \citep{bender-koller-2020-climbing}. Large language models (LLMs) show remarkable results on \textit{vertical thinking} tasks that require sequential logical inference \citep{liu2019roberta} but there have been relatively few studies done on \textit{lateral thinking} puzzles\textemdash tasks that require more creative, “outside the box" problem-solving processes. As larger LLMs with the ability to memorize large corpora \citep{hartmann2023sok} are developed, lateral thinking tasks become an increasingly important benchmark for analyzing and evaluating their reasoning capacities. The \textsc{BrainTeaser} shared task \citep{jiang-etal-2023-brainteaser}\citep{jiang-ilievski-ma:2024:SemEval2024} is designed to elicit and evaluate lateral thinking through two English-language subtasks, using sentence puzzles and word puzzles respectively.

In this paper, we propose a novel method for optimizing chain-of-thought (CoT) prompting \citep{wei2023chainofthought} on the \verb|GPT-4| model which we use to tackle the sentence puzzle subtask. Our system iteratively optimizes CoT prompting by systematically evaluating input data and model output using human performance as a benchmark. We identify question types that are difficult for humans, informing the next iteration of prompt engineering. Not only does this process optimize CoT prompting for a specific task, our system also provides insights for improving future data collection and synthesis.

Our main contribution is the novel approach for identifying reasoning challenges to optimize prompting. For the sentence-based task, we develop a prompt engineering method which requires the model to reason over all answer choices and provide explanations for both correct and incorrect choices. In doing so, the model is more likely to refute choices that are semantically related to the question but logically incorrect. Our methodology significantly improves performance for adversarial datasets and achieves more consistent results, which suggests that the model relies less on memorization when using these CoT prompts. 

As part of our evaluation of the data, we also identify several questions in the adversarial datasets that are difficult to solve due to having multiple logical options or are unanswerable with the provided premises. By combining model reasoning with human evaluation, we can quickly identify and evaluate problematic questions. This process can further explain model performance and provide guidance for future data collection/generation.
\section{Background}
Question Answering (QA) is a well-established task in natural language processing with broad applications both in academia and in industry \citep{hirschman_gaizauskas_2001}. Recent work such as CommonSenseQA (CSQA) \citep{DBLP:journals/corr/abs-1811-00937} and StrategyQA \citep{geva2021did} focus on reasoning questions that require logical inference in the form of vertical thinking. \textsc{BrainTeaser} questions instead require lateral thinking to answer, much like questions in the traditional "brainteaser" style\citep{jiang-etal-2023-brainteaser}\citep{jiang-ilievski-ma:2024:SemEval2024}:

\texttt{{\small \begin{itemize}
    \item[]\textbf{Base: }Samuel was out for a walk when it started to rain. He did not have an umbrella and he wasn't wearing a hat. His clothes were soaked, yet not a single hair on his head got wet. How could this happen?
    \begin{enumerate}
        \item His hair is dyed.
        \item \textbf{This man is bald.}
        \item This man walk upside down to avoid rain.
        \item None of above.
    \end{enumerate}
    \item[]\textbf{SR: }Rain began to fall as Samuel was taking a stroll. He wasn't wearing a hat, and he didn't have an umbrella. Even though his clothes were completely drenched, not a single hair on his head was moist. How is this even possible?
    \begin{enumerate}
        \item This man walk upside down to avoid rain.
        \item His hair is dyed.
        \item \textbf{This man is bald.}
        \item None of above.
    \end{enumerate}
        \item[]\textbf{CR: }Tom is a clean freak but he never dries his hair after a shower. How is this possible?
    \begin{enumerate}
        \item His hair is dyed.
        \item He tries to stand upside down during shower to avoid rain.
        \item \textbf{This man is bald.}
        \item None of above.
    \end{enumerate}
\end{itemize}}}

The data for this subtask is drawn from online English-language riddles and brainteasers, with incorrect choices created by handpicking entailments generated by \verb|COMET| \citep{bosselut-etal-2019-comet} using incorrect premises. Each question has three unique answers, as well as a shared fourth option, "None of above". To counter memorization from LLMs trained on web crawls, the task authors generated two synthetic datasets using \textit{semantic reconstruction} (SR) and \textit{context reconstruction} (CR). The SR dataset rephrases the original question without changing the answer or premises while the CR dataset changes the situational context without changing the misleading premise. The SR dataset was generated with an open-source rephrasing tool while the CR dataset was generated using \verb|GPT-4|; both sets were manually refined by human annotators. In total, 208 question/answer pairs were sampled for the base set resulting in 624 questions after SR and CR augmentation. The training set was split with 81.25\% of the data with the same base/SR/CR questions kept together in the split. 

Although the task is designed to elicit lateral thinking, we consider an alternative understanding of the task by thinking of the questions as \textit{noisy}. Questions are loaded with irrelevant, contradictory, or misleading information to distract the respondent. Since transformers generally learn meaning by scoring tokens across the sentence or sentence pair, they are biased against long-tail knowledge \citep{li2023search}, which is knowledge that occurs infrequently in the training set.

Brainteasers, by their nature, rely on the unconventional interpretation of the question to stump the respondent. This same property can trick the model into selecting a semantically similar answer choice that is logically incorrect. Chain-of-Thought (CoT) prompt engineering \citep{wei2023chainofthought} is a recent method that has been shown to not only improve outcomes on similar problems, but also to provide an interpretable window for human review. CoT prompts provide example questions with related reasoning to the model, which induces the model to provide reasoning for a given answer in the output. We utilize both of these properties to introduce an iterative method that optimizes CoT prompting for a given task.
\section{System Overview} 
We propose an iterative system for optimizing the CoT prompt engineering process:
\begin{enumerate}
    \item Randomly sample the training data and naively engineer CoT prompts.
    \item Identify distinct categories in output reasoning to partition training data.
    \item Perform independent human evaluation to isolate specific challenges in each category.
    \item Use findings to inform the development of new CoT prompts.
    \item Optionally, identify gaps in the data for future data collection/synthesis.
\end{enumerate}

Each step of our process is iterative, although independent human evaluation should only be performed when novel problem categories are identified in the model reasoning. Once the human benchmark has been incorporated into the prompt engineering process, future iterations mainly rely on evaluating model outputs for gaps in logic. All evaluation steps can provide powerful insights into the dataset itself to inform future dataset creation, and is particularly useful for real-world applications where data selection is an open-ended problem. By identifying gaps and problems in the data, more representative data can be collected to improve model performance on the given application. This can be thought of as a backpropagating human and model outputs back to the prompt engineering and data curation steps.

\subsection{Naive Chain-of-Thought Prompting}
In the first step, we randomly sample the training data to generate naive CoT prompts to use on the test set. This step eliminates a large portion of the dataset that naive CoT prompting already solves while also providing outputs with interpretable windows for identifying problem questions and corresponding failure in logic. For example, the topics (mathematics, physics, law, etc.) identified by the task authors in the training data were found to have minimal impact on model accuracy. Instead, we found that the construction type impacted model performance above all. By focusing on the model outputs, we determine that the type of reconstruction (base, SR, and CR) have the highest impact on model performance and require human analysis.

For the first round of CoT prompting, we randomly select 8 samples from the training set and generate the Naive CoT-Base prompt set based on the logical premises of the questions (all naive prompts in Appendices \ref{naive-cot}):

\texttt{{\small Naive CoT Example Prompt:\\
Question: A horse is tied to a five-meter rope in front of an old saloon. Ten meters behind the horse is a bale of hay. Without breaking his rope, the horse is able to eat the hay whenever he chooses. How is this possible?\\
Choices: \\
0 = The rope stretches proportionally, providing the extra length needed for the horse to reach the hay ten meters away.;\\
1 = The rope is not tied to anything else.;\\
2 = The walls of the saloon retract or collapse inwards, creating more space for the horse to reach the hay.;\\
3 = None of above.;\\
Response: That the rope is not tied to anything else is the simplest choice. The horse can reach the hay whenever he chooses. The answer is 1}}
\subsection{Human Evaluation Step}
Based on model performance using the naive CoT prompts, we separate the test set along base, SR, and CR lines for human testing. When prompting \verb|GPT-4|, each question is independently shown to the model with no retention in between. Humans are not under similar constraints and can easily identify reconstructions, especially the SR set which share the same answer choices with the base set. For accurate comparison, we select different participants to answer each dataset.

We selected 3 participants for each dataset in order to collect robust results while still maintaining consensus. All participants are graduate students with native proficiency in English, and surveys were completed using Google Forms with the same instructions, randomized question order, and constant answer choice order. An additional option "Unsure" was provided to uncover difficult questions which was counted as "None of above" for testing purposes. 

We analyze accuracy along four metrics: mean, minimum score, maximum score, and consensus score. The minimum score is counted only when all participants answer correctly whereas the maximum score is counted when any participant answers correctly. The consensus score uses the answer selected by 2/3 participants; 4 questions had no consensus and were marked as incorrect. Through this process, we identified common errors for humans and models in each dataset which informed the second iteration of CoT prompting. 

\begin{table*}[!h]
\centering
\begin{tabular}{l|ccc|cc|c}
\noalign{\global\arrayrulewidth0.9pt}
\hline
\noalign{\global\arrayrulewidth0.4pt}
\multirow{2}*{\textbf{System}} & \multicolumn{3}{c|}{\textbf{Instance Based}} & \multicolumn{2}{c|}{\textbf{Group Based}} & \multirow{2}*{\textbf{Overall}}\\
 & \textbf{Base} & \textbf{SR} & \textbf{CR} & \textbf{Base\&SR} & \textbf{Adversarial} & \\
\noalign{\global\arrayrulewidth0.9pt}
\hline
\noalign{\global\arrayrulewidth0.4pt}
abdelhak & 100 & 100 & 95.0 & 100 & 95.0 & 98.3\\
\hline
Human \citep{jiang-etal-2023-brainteaser} & 90.7 & 90.7 & 94.4 & 90.7 & 88.9 & 92.0\\
Human Consensus (Ours) & 90.0 & 90.0 & 67.5 & 80.0 & 55.0 & 82.5\\
\hline
\verb|GPT-4| Zero Shot & 87.5 & 72.5 & 70.0 & 72.5 & 60.0 & 76.7\\
\verb|GPT-4| Multi Shot Base & 92.5 & \textbf{90.0} & 80.0 & \textbf{87.5} & 70.0 & 87.5\\
\verb|GPT-4| Multi Shot Mix & \textbf{95.0} & \textbf{90.0} & \textbf{85.0} & \textbf{87.5} & \textbf{80.0} & \textbf{90.0}\\
\hline
\verb|GPT-4| Naive CoT-Base & \textbf{95.0} & \textbf{87.5} & 75.0 & 85.0 & 65.0 & 85.8 \\
\verb|GPT-4| Naive CoT-Mix & 92.5 & \textbf{87.5} & \textbf{82.5} & \textbf{87.5} & \textbf{75.0} & \textbf{87.5} \\
\hline
\verb|GPT-4| New CoT-Base & \textbf{97.5} & 85.0 & 80.0 & 85.0 & 70.0 & 87.5\\
\verb|GPT-4| New CoT-SR & 90.0 & 90.0 & 75.0 & 85.0 & 67.5 & 85.0\\
\verb|GPT-4| New CoT-CR & 92.5 & 90.0 & 77.5 & 87.5 & 67.5 & 86.7\\
\verb|GPT-4| New CoT-Mix & 95.0 & \textbf{92.5} & \textbf{82.5} & \textbf{92.5} & \textbf{77.5} & \textbf{90.0}\\
\noalign{\global\arrayrulewidth0.9pt}
\hline
\noalign{\global\arrayrulewidth0.4pt}
\end{tabular}
\caption{Accuracy of each model; 
\textbf{Base}, \textbf{SR}, and \textbf{CR} are scored on individual datasets, 
\textbf{Base\&SR} only counts if both the base and SR versions are correct for a given question,
\textbf{Adversarial} only counts if all three versions are correct for a given question, and 
\textbf{Overall} counts base, SR, and CR separately.
}
\label{results}
\end{table*}
\subsection{Iterated Chain-of-Thought Prompting}
Besides the tricky CR questions uncovered in the human evaluation step, we also identified that our naive sample was overly weighted on base questions, which potentially serves to reinforce model memorization. To address both issues, we develop the CoT-Mix set, a new set of 8 prompts weighted towards SR and CR questions and tailored towards disproving incorrect answer choices. Since our human benchmark performed particularly poorly on CR questions, we also separately created CoT prompts comprised entirely of base, SR, and CR sets for further comparison. All prompt sets can be found in the Appendices \ref{new-cot}.

\texttt{{\small Iterated Chain-of-Thought Prompt:\\
Question: A horse is tied to a five-meter rope in front of an old saloon. Ten meters behind the horse is a bale of hay. Without breaking his rope, the horse is able to eat the hay whenever he chooses. How is this possible\\
Choices:\\
0 = The rope stretches proportionally, providing the extra length needed for the horse to reach the hay ten meters away.;\\
1 = The rope is not tied to anything else.;\\
2 = The walls of the saloon retract or collapse inwards, creating more space for the horse to reach the hay.;\\
3 = None of above.;\\
Response: Rope generally cannot stretch, and if it could stretch the length would be variable. If the walls collapse, the horse would be further from the hay. The rope not being tied to anything else is the simplest answer. The answer is 1}}

\section{Experimental Setup}
All experiments were performed using \verb|GPT-4| via the OpenAI API\footnote{\url{https://platform.openai.com/}} with the same system prefix. Each question was called separately with the respective prefix. Initial experiments limited token count to 1 in order to force the model to output integer labels but the restriction was removed due to poor performance. Output text was logged and labels were extracted deterministically; labels that could not be extracted this way were reviewed and manually entered. Results shown in this paper were from API calls between 2024/01/15-2024/02/17; since OpenAI models are updated regularly replication results may differ.

The input data was first preprocessed by removing extra spaces, lines, punctuation, and spelling and grammatical errors. Due to the variety of sentence-based problems in the data, some answer choices were multiple sentences long while others were single words; this discrepancy could potentially affect model performance. Since semicolons do not occur in the data at all, they were selected as separators between answer choices for \verb|GPT-4| prompting to mitigate this issue. System prompts can be found in Appendix \ref{system}.
\section{Results}
In this section, we compare results across four categories: human performance, zero/multi-shot performance, naive CoT prompt performance, and the iterated CoT prompt performance. For completion, we also provide the top competition result (abdelhak). Our official submission for the shared task leaderboard used the \verb|GPT-4| New CoT-Base prompts. With 31 participants, our results scored 2nd overall on the base data, 7th on SR, CR, and Base\&SR accuracy, 9th on adversarial accuracy, and 9th overall.
\subsection{Quantitative Benchmarks}
As expected, the new CoT prompts show significant improvements along all metrics compared to their naive counterparts. Since the adversarial datasets are designed specifically to counteract model memorization, the gains made in the group-based metrics between the Naive and New CoT-Mix prompts in particular demonstrates the effectiveness of our system. The CoT-Base sets, on the other hand, likely still suffer from overweighting on the original questions crawled from online sources. The fact that multi-shot prompting outperforms the Naive CoT prompts further supports this argument. Despite not receiving any guidance on logic, the model is still able to achieve strong results on the task. However, once we introduce the idea of disproving incorrect answers, the model is once again able to make gains in performance.

Interestingly, the SR and CR prompts did not show significant improvements compared to the base set and even performed worse on the CR questions. This could potentially be due to the fact that the model performs worse on these question types overall. Without base questions to provide a foundation, the model is unable to generate the most robust reasoning. While this finding provides grounds for further exploration, the results from the New CoT-Mix prompting shows consistent improvements across the board, suggesting that there are commonalities within each adversarial dataset that can be identified by the model. 
\subsection{Human Performance Evaluation}
\begin{table}[!h]
\centering
\begin{tabular}{lccccc}
\hline
\textbf{Dataset} & \textbf{Mean} & \textbf{Min.} & \textbf{Con.} & \textbf{Max.}\\
\hline
Base & 84.2 & 65.0 & 87.5 & 100.0\\
SR & 85.8 & 70.0 & 90.0 & 97.5\\
CR & 60.0 & 30.0 & 65.0 & 80.0\\
\hline
\end{tabular}
\caption{Human participant accuracy for each dataset; 
\textbf{Min.} is only counted when all three participants answer correctly, 
\textbf{Con.} is counted when 2/3 agree on an answer,
and \textbf{Max.} is counted when any participant scores correctly.}
\label{human results}
\end{table}
When evaluating human performance, we find that there were significantly higher rates of "Unsure" answers in the CR set which contributes to the lower overall score. Along with other observations, the greater rate of "Unsure" answers supports the idea that the CR questions are more difficult to reason over. However, there were no cases where all respondents selected "Unsure" for the same question, making this metric a weak indicator of problematic questions. One possible explanation suggests that the wording for the answers in the CR set are less well-formed than the other sets, leading to greater confusion among respondents. 

For human performance, none of our CoT results were able to beat the task paper human benchmark on the CR dataset, although our iterated CoT results did surpass the human benchmark on SR. Our human results only drastically differ on the CR set with those found in the task paper, which could be influenced by several factors. A major factor is the fact that we use independent human annotators for each of the datasets, meaning that those working on the CR set have no reference to the original question. For questions with multiple valid answers or unclear logic, our annotators would not be able to reference the base and SR versions for clues. Out of 14 incorrectly answered CR questions, we identify 5 such questions.

Specifically, we discovered several questions in the CR set with multiple logical responses. In the example below, the correct answer is "two and a half hours" with the assumption that each stop is meant to take half an hour. Ignoring that the premise itself is nonsensical (the driver would have simply been told to stop for the full two and a half hours), nowhere is it stated that the driver takes half an hour for each break. As a result, both (2.) and (4.) are viable candidates depending on reasoning. The rest of the problem questions can be seen in Appendix \ref{cr}.
\newline
\texttt{{\small Example Question: A driver is told to make a stop every half an hour for the engine to cool down, for five times. How long do the stops take?\\
Choices:\\
1. Three hours.\\
\textbf{2. Two and a half hours.}\\
3. Two hours.\\
4. None of above.\\}}

\section{Conclusions}
In this paper, we demonstrate a novel system for optimizing chain-of-thought prompt engineering using human evaluation. While Naive CoT prompts performed similarly to multi-shot prompting without guidance on reasoning, later iterations were able to approach state-of-the-art performance. Significant improvement on the SR, CR, and group-based metrics were shown on the test data, supporting the adoption for this method of prompt engineering. This system further provides guidance on identifying key problem areas in the dataset, specifically with regards to the generation of context reconstruction questions. This form of evaluation serves to inform decisions for improving dataset quality. For future work, we plan to implement different techniques to create synthetic data and perform the same evaluation across an open-ended dataset. 

\bibliography{anthology, custom}

\appendix

\section{Naive CoT Prompts}
\label{naive-cot}
\subsection{Naive Base Prompts}
\begin{enumerate}
    \item Question: Samuel was out for a walk when it started to rain. He did not have an umbrella and he wasn't wearing a hat. His clothes were soaked, yet not a single hair on his head got wet. How could this happen?\\
    Choices:\\
    0 = His hair is dyed.;\\
    1 = This man is bald.;\\
    2 = This man walk upside down to avoid rain.;\\
    3 = None of above.;\\
    Response: Samuel got wet from the rain, but no hair on his head got wet. If Samuel dyed his hair it would still be wet. Samuel is bald, so his hair didn't get wet. The answer is 1
    \item Question: A horse is tied to a five-meter rope in front of an old saloon. Ten meters behind the horse is a bale of hay. Without breaking his rope, the horse is able to eat the hay whenever he chooses. How is this possible?\\
    Choices: \\
    0 = The rope stretches proportionally, providing the extra length needed for the horse to reach the hay ten meters away.; \\
    1 = The rope is not tied to anything else.; \\
    2 = The walls of the saloon retract or collapse inwards, creating more space for the horse to reach the hay.; \\
    3 = None of above.;\\
    Response: That the rope is not tied to anything else is the simplest choice. The horse can reach the hay whenever he chooses. The answer is 1
    \item Question: A woman who lives in new york legally married three men, she did not get divorce, get an enrollment, or legally seperate. How is this possible?\\
    Choices: \\
    0 = The woman is not a good person.; \\
    1 = His husband is not a good husband.; \\
    2 = She is a minister.; \\
    3 = None of above.;\\
    Response: In new york it is not legal to be married to more than one person at a time. If the woman did not get divorced or legally separated, she cannot be legally married to more than one person. Ministers perform marriages, they do not get married. The answer is 2
    \item Question: Why is the value of 1968 pennies higher than 1967 pennies?\\
    Choices: \\
    0 = Old money weight more than new money.; \\
    1 = Old money is dirtier than new money.; \\
    2 = Because there is one more penny in 1968 pennies than in 1967 pennies.; \\
    3 = None of above.;\\
    Response: 1968 and 1967 refer to the number of pennies. The value of 1968 pennies is higher than 1967 pennies because there is one more penny in 1968 pennies than in 1967 pennies. The answer is 2
    \item Question: Not a single parent objected when the teacher spanked every child in the class. How come?\\
    Choices: \\
    0 = The teacher had informed all the parents in advance about the unique disciplinary approach.; \\
    1 = The teacher had informed all the children in advance about the unique disciplinary approach.; \\
    2 = The teacher was in an orphanage school.; \\
    3 = None of above.;\\
    Response: Even if the teacher had informed all the parents or all the students in advance about the unique disciplinary approach, it is unlikely that nobody would object. If the teacher is in an orphange school, there would be no parents to object. The answer is 2
    \item Question: Twenty-seven ducks are going to the pond. Five of them got lost, thirteen of them are staying home, and nine of them are at the pond. Where are the rest of them?\\
    Choices: \\
    0 = Home.; \\
    1 = The way to the pond.; \\
    2 = Pond.; \\
    3 = None of above.;\\
    Response: Twenty-seven ducks are going to the pond. Five of them are lost, 27 - 5 = 22. Thirteen of them are staying home, 22 - 13 = 9. Nine of them are at the pond, 9 - 9 = 0. There are no more ducks so the question is invalid. The answer is 3
    \item Question: How many birth days does the average person have?\\
    Choices: \\
    0 = People may celebrate their birthdays annually, so it depends on their life span.; \\
    1 = They technically only have one birth day in their lifetime.; \\
    2 = It can be zero as some people are too busy to celebrating their birthdays.; \\
    3 = None of above.;\\
    Response: Everyone is born only once so technically the average person has one birthday in their lifetime. The answer is 1
    \item Question: The more you take, the more you leave behind\\
    Choices: \\
    0 = Love.; \\
    1 = Footsteps.; \\
    2 = Money.; \\
    3 = None of above.;\\
    Response: Footsteps. The more you walk, the more footsteps you leave behind. The answer is 1
\end{enumerate}
\subsection{Naive Mixed Prompts}
\begin{enumerate}
    \item Question: A horse is tied to a five-meter rope in front of an old saloon. Ten meters behind the horse is a bale of hay. Without breaking his rope, the horse is able to eat the hay whenever he chooses. How is this possible?\\
    Choices: \\
    0 = The rope stretches proportionally, providing the extra length needed for the horse to reach the hay ten meters away.; \\
    1 = The rope is not tied to anything else.; \\
    2 = The walls of the saloon retract or collapse inwards, creating more space for the horse to reach the hay.; \\
    3 = None of above.;\\
    Response: Rope generally cannot stretch, and if it could stretch the length would be variable. If the walls collapse, the horse would be further from the hay. The rope not being tied to anything else is the simplest answer. The answer is 1
    \item Question: A woman who lives in new york legally married three men, she did not get divorce, get an enrollment, or legally seperate. How is this possible?\\
    Choices: \\
    0 = The woman is not a good person.; \\
    1 = His husband is not a good husband.; \\
    2 = She is a minister.; \\
    3 = None of above.;\\
    Response: In new york it is not legal to be married to more than one person at a time, even if her husband is bad. If the woman did not get divorced or legally separated, she cannot be legally married to more than one person. Ministers perform marriages, they do not get married. The answer is 2
    \item Question: Brad began by entering the office tower's 22nd story through the filthy common window. He slid the window open and leapt through it after becoming depressed. Outside the building, there was a cliff-like drop to the ground. Amazingly, he came to rest entirely unharmed. How could he have survived the fall if there was nothing to slow or cushion his descent?\\
    Choices: \\
    0 = Brad was so sick and tired of window washing, he opened the window and jumped inside.; \\
    1 = The ground outside the building is wet.; \\
    2 = Consistent exercise has made him a very strong man.; \\
    3 = None of above.;\\
    Response: Even very strong men would not survive a fall from a 22-story building. The ground outside the building being wet would not cushion his fall. Brad was so sick and tired of window washing, he opened the window and jumped inside. The answer is 0
    \item Question: Danny had just passed under an overpass in his semi when he abruptly came to a stop. Danny accidentally drove under the overpass that was only just tall enough for his truck because he wasn't paying enough attention. He was unable to move forward or backward in the semi due to how tightly it was wedged. When another tracker passed by, he was told how simple it would be to remove the semi from underneath the bridge. What did he recommend?\\
    Choices: \\
    0 = He told Danny to left the bridge.; \\
    1 = He told Danny to overturn the track.; \\
    2 = He told Danny to let some air out of his tires.; \\
    3 = None of above.;\\
    Response: The bridge would not be lifted. Overturning the truck would not help. Letting some air out of his tires would help. The answer is 2
    \item Question: I excavate little caves and keep my gold and silver there. I also create gold crowns and silver bridges. They are the tiniest things you can imagine. Everyone will eventually require my assistance, but many people are reluctant to accept it. Why?\\
    Choices: \\
    0 = I am a dentist.; \\
    1 = I am a thief.; \\
    2 = I am a miner.; \\
    3 = None of above.;\\
    Response: Thieves and miners do not create gold crowns or bridges. Tiny crowns and bridges are dental work. People don't like going to the dentist. The answer is 0
    \item Question: Each of the 30 participants in the masquerade had to wear a unique hat to distinguish themselves from one another. The host, however, only tallied 29 when he counted the number of hats to determine attendance. All 30 persons had signed their names on the spreadsheet, which confused him. He repeated the count. There are still 29. How is that even doable?\\
    Choices: \\
    0 = One person had a pretty beautiful hat.; \\
    1 = The host had a hat himself and he forget to count it.; \\
    2 = One person had moved away from the group when the host was counting.; \\
    3 = None of above.;\\
    Response: If one person had a pretty beautiful hat, then the host would still count 30 hats. If one person had moved away from the group when the host was counting, then the host would still count 30 hats. The host had a hat himself and he forget to count it. The answer is 1
    \item Question: A certain kind of animal has parents but no children, lives happily but can not give birth to offspring. This kind of animal has existed on earth for a long history. How is that possible?\\
    Choices: \\
    0 = The animal is the dinosaur. Millions of years ago, dinosaurs roamed the Earth, but eventually, they became extinct due to various factors such as climate change or asteroid impact.; \\
    1 = The animal is the butterfly. During the pupal stage of their life cycle, butterflies are inside a protective casing, and it may appear as if they have completely disappeared or been wiped out.; \\
    2 = The animal is the Mule. Since all Mules are born sterile, you can only get a Mule by crossing a donkey with a horse.; \\
    3 = None of above.;\\
    Response: Dinosaurs are no longer alive, so they cannot live happily. Butterflies can give birth to offspring, so they are not the answer. Mules are born sterile, so they cannot give birth to offspring. The answer is 2
    \item Question: Six women were hiking on a trail when a sudden snowfall began. Five of the women hurried along, but the sixth did not. Still, they all reached the cabin at the same time, and all but the sixth were cold. How is that possible?\\
    Choices: \\
    0 = The 6th woman beated them at the end.; \\
    1 = The sixth woman was in a photograph that one of the others was carrying.; \\
    2 = The 6th woman walked slowly to dodge the snow.; \\
    3 = None of above.;\\
    Response: If the sixth woman beat them or walked slowly to dodge the snow she would not have reached the cabin at the same time. If the sixth woman was in a photograph that one of the others was carrying, then she would not be cold and would not have to hurry. The answer is 1
\end{enumerate}

\section{New CoT Prompts}
\label{new-cot}
\subsection{New Base Prompts}
\begin{enumerate}
    \item Question: A couple is having a disagreement over the man's error. The man kept apologizing and pleading with the woman to allow him to see her directly. The woman was still upset. Thus, she wouldn't agree. The couple, though, was positioned on the same mat. How is that even doable?\\
    Choices: \\
    0 = The mat was folded in multiple times and end up with a special shape.; \\
    1 = The woman kept turning her back to the man.; \\
    2 = The couple was standing on two sides of the girl's apartment door, which had a rug under it.; \\
    3 = None of above.;\\
    Response: The couple are standing on the same mat but cannot see each other, which suggests that there is a barrier between them. Regardless of how the mat was folded, it would not prevent the couple from seeing each other. Even if the woman turned her back to the man he would still see her. Therefore, the couple was standing on two sides of a door, which had a rug under it. The answer is 2
    \item Question: Eight people were sitting under a large tree. Suddenly, a gust of wind blows, yet none of them got hit by any falling leaves. How is this possible?\\
    Choices: \\
    0 = It was winter and the tree doesn't have any leaves.; \\
    1 = People were camping under the tree.; \\
    2 = The wind blows heavily.; \\
    3 = None of above.;\\
    Response: If the people were camping under the tree, they would still be hit by the falling leaves. The wind blowing heavily would cause the leaves to fall. If it was winter and the tree doesn't have any leaves, then the people would not be hit by any falling leaves. The answer is 0
    \item Question: The ship was in the central Pacific Ocean. The ship suddenly began to sink without being crushed. However, each team was still preoccupied with its own tasks, so none was threatening. Why?\\
    Choices: \\
    0 = An underwater earthquake caused a rapid drop in the water level and resulted in the loss of buoyancy for the ship.; \\
    1 = It was a Submarine.; \\
    2 = There are too many fish around the ship.; \\
    3 = None of above.;\\
    Response: An earthquake would cause people to feel threatened, and fish around the ship would not matter. If the ship were a submarine, it would be designed to sink. The answer is 1
    \item Question: Two mothers and two daughters were asking for new state IDs, but the agent only gave out three forms and instructed them on how to fill them out. Why?\\
    Choices: \\
    0 = They are one daughters, one mother and one grandmother.; \\
    1 = Two girls filled the from together.; \\
    2 = One mother is too old to apply for new IDs.; \\
    3 = None of above.;\\
    Response: The agent only gave out three forms and instructed them on how to fill them out, which suggests that there are only three people. If they are one daughter, one mother, and one grandmother, then there would be three people. The answer is 0
    \item Question: Every night, a man would sleep with a light on, as bright as the sun, dazzling the neighbors. But why did his neighbors never complain?\\
    Choices: \\
    0 = All his neighbors are blind.; \\
    1 = The man would wake up early.; \\
    2 = He lived in a light house.; \\
    3 = None of above.;\\
    Response: If all his neighbors are blind, they would not be dazzled by the light. If the man lived in a light house then the light would be expected. The answer is 2 
    \item Question: Each of the 30 participants in the masquerade had to wear a unique hat to distinguish themselves from one another. The host, however, only tallied 29 when he counted the number of hats to determine attendance. All 30 persons had signed their names on the spreadsheet, which confused him. He repeated the count. There are still 29. How is that even doable?\\
    Choices: \\
    0 = One person had a pretty beautiful hat.; \\
    1 = The host had a hat himself and he forget to count it.; \\
    2 = One person had moved away from the group when the host was counting.; \\
    3 = None of above.;\\
    Response: If one person had a pretty beautiful hat, then the host would still count 30 hats. If one person had moved away from the group when the host was counting, then the host would still count 30 hats. The host had a hat himself and he forget to count it. The answer is 1
    \item Question: A certain kind of animal has parents but no children, lives happily but can not give birth to offspring. This kind of animal has existed on earth for a long history. How is that possible?\\
    Choices: \\
    0 = The animal is the dinosaur. Millions of years ago, dinosaurs roamed the Earth, but eventually, they became extinct due to various factors such as climate change or asteroid impact.; \\
    1 = The animal is the butterfly. During the pupal stage of their life cycle, butterflies are inside a protective casing, and it may appear as if they have completely disappeared or been wiped out.; \\
    2 = The animal is the Mule. Since all Mules are born sterile, you can only get a Mule by crossing a donkey with a horse.; \\
    3 = None of above.;\\
    Response: Dinosaurs are no longer alive, so they cannot live happily. Butterflies can give birth to offspring, so they are not the answer. Mules are born sterile, so they cannot give birth to offspring. The answer is 2
    \item Question: Six women were hiking on a trail when a sudden snowfall began. Five of the women hurried along, but the sixth did not. Still, they all reached the cabin at the same time, and all but the sixth were cold. How is that possible?\\
    Choices: \\
    0 = The 6th woman beated them at the end.; \\
    1 = The sixth woman was in a photograph that one of the others was carrying.; \\
    2 = The 6th woman walked slowly to dodge the snow.; \\
    3 = None of above.;\\
    Response: If the sixth woman beat them or walked slowly to dodge the snow she would not have reached the cabin at the same time. If the sixth woman was in a photograph that one of the others was carrying, then she would not be cold and would not have to hurry. The answer is 1
\end{enumerate}
\subsection{New Mixed Prompts}
\begin{enumerate}
    \item Question: A horse is tied to a five-meter rope in front of an old saloon. Ten meters behind the horse is a bale of hay. Without breaking his rope, the horse is able to eat the hay whenever he chooses. How is this possible?
    Choices: \\
    0 = The rope stretches proportionally, providing the extra length needed for the horse to reach the hay ten meters away.; \\
    1 = The rope is not tied to anything else.; \\
    2 = The walls of the saloon retract or collapse inwards, creating more space for the horse to reach the hay.; \\
    3 = None of above.;\\
    Response: That the rope is not tied to anything else is the simplest choice. The horse can reach the hay whenever he chooses. The answer is 1
    \item Question: A woman who lives in new york legally married three men, she did not get divorce, get an enrollment, or legally seperate. How is this possible?\\
    Choices: \\
    0 = The woman is not a good person.; \\
    1 = His husband is not a good husband.; \\
    2 = She is a minister.; \\
    3 = None of above.;\\
    Response: In new york it is not legal to be married to more than one person at a time. Ministers perform marriages, they do not get married. The answer is 2\\
    \item Question: Brad began by entering the office tower's 22nd story through the filthy common window. He slid the window open and leapt through it after becoming depressed. Outside the building, there was a cliff-like drop to the ground. Amazingly, he came to rest entirely unharmed. How could he have survived the fall if there was nothing to slow or cushion his descent?\\
    Choices: \\
    0 = Brad was so sick and tired of window washing, he opened the window and jumped inside.; \\
    1 = The ground outside the building is wet.; \\
    2 = Consistent exercise has made him a very strong man.; \\
    3 = None of above.;\\
    Response: A person cannot jump out of a 22 story building without injury. Brad must have jumped into the building. The answer is 0
    \item Question: Danny had just passed under an overpass in his semi when he abruptly came to a stop. Danny accidentally drove under the overpass that was only just tall enough for his truck because he wasn't paying enough attention. He was unable to move forward or backward in the semi due to how tightly it was wedged. When another tracker passed by, he was told how simple it would be to remove the semi from underneath the bridge. What did he recommend?\\
    Choices: \\
    0 = He told Danny to left the bridge.; \\
    1 = He told Danny to overturn the track.; \\
    2 = He told Danny to let some air out of his tires.; \\
    3 = None of above.;\\
    Response: Danny has to reduce the height of his truck to pass the bridge. Letting the air out of his tires reduces the truck's height. The answer is 2
    \item Question: I excavate little caves and keep my gold and silver there. I also create gold crowns and silver bridges. They are the tiniest things you can imagine. Everyone will eventually require my assistance, but many people are reluctant to accept it. Why?\\
    Choices: \\
    0 = I am a dentist.; \\
    1 = I am a thief.; \\
    2 = I am a miner.; \\
    3 = None of above.;\\
    Response: Tiny crowns and bridges are dental work. People don't like going to the dentist. The answer is 0
    \item Question: Each of the 30 participants in the masquerade had to wear a unique hat to distinguish themselves from one another. The host, however, only tallied 29 when he counted the number of hats to determine attendance. All 30 persons had signed their names on the spreadsheet, which confused him. He repeated the count. There are still 29. How is that even doable?\\
    Choices: \\
    0 = One person had a pretty beautiful hat.; \\
    1 = The host had a hat himself and he forget to count it.; \\
    2 = One person had moved away from the group when the host was counting.; \\
    3 = None of above.;\\
    Response: The host is the one counting the hats. The host had a hat himself and he forget to count it. The answer is 1
    \item Question: A certain kind of animal has parents but no children, lives happily but can not give birth to offspring. This kind of animal has existed on earth for a long history. How is that possible?\\
    Choices: \\
    0 = The animal is the dinosaur. Millions of years ago, dinosaurs roamed the Earth, but eventually, they became extinct due to various factors such as climate change or asteroid impact.; \\
    1 = The animal is the butterfly. During the pupal stage of their life cycle, butterflies are inside a protective casing, and it may appear as if they have completely disappeared or been wiped out.; \\
    2 = The animal is the Mule. Since all Mules are born sterile, you can only get a Mule by crossing a donkey with a horse.; \\
    3 = None of above.;\\
    Response: The animal is still alive, has parents, but cannot have children. Mules are born sterile, so they cannot give birth to offspring. The answer is 2
    \item Question: Six women were hiking on a trail when a sudden snowfall began. Five of the women hurried along, but the sixth did not. Still, they all reached the cabin at the same time, and all but the sixth were cold. How is that possible?\\
    Choices: \\
    0 = The 6th woman beated them at the end.; \\
    1 = The sixth woman was in a photograph that one of the others was carrying.; \\
    2 = The 6th woman walked slowly to dodge the snow.; \\
    3 = None of above.;\\
    Response: All the women hurried except the sixth and were cold. The sixth woman is not physically present. She must be in a photograph one of the others was carrying. The answer is 1
\end{enumerate}

\section{System and User Prompts}
\label{system}

"role": "system", 
"content": "You are a Question Answering Model that answers questions by finding logical entailments between the question and answer choices."

\section{Problematic CR Questions}
\label{cr}
\begin{enumerate}
    \item SP-120\_CR: Mark was in a playground where somebody noticed a great player playing and with the announcements, gathered a lot of people. There were many great players from basketball, volleyball, football, and even swimmers, But Mark directly went to the footballer and took a photo with him. How did he know who was the person that people got excited for in the first place?\\
    \textbf{"Since the playground was a football playground and the other players could've not been playing in the playground at the time of the announcement.",}\\
    "Since Mark was a crazy fan of football, only a football player can be considered as great player in his mind.",\\
    "Since Mark stood closest to the football players, he only focused on football players and didn't notice others.",\\
    "None of above."\\
    \item SP-30\_CR: Why do old people consume more food than young people.\\
    "Older adults may have specific dietary requirements to address age-related issues",\\
    "Older people require increased nutrient intake to support overall health and well-being.",\\
    \textbf{"Because older people live longer.",}\\
    "None of above."\\
    \item SP-184\_CR: Five people were at a football match, and a sudden shower started. The four that rushed to take cover still got soaked, but the one who didn't move stayed completely dry. Why didn't he get wet?\\
    "The man is an excellent football player that can avoid rain in high speed."\\
    "The man was lucky enough to avoid all the rain.\\
    \textbf{"He was a photograph, the other people were there to honor a former player.",}\\
    "None of above."\\
    \item SP-166\_CR: A farmer has 11 sheep. Half of them are white. How is this possible?\\
    "One sheep is regarded as both white and other colors same time.",\\
    "A famer raises his sheep in both white way and another way.",\\
    \textbf{"They are all white.",}\\
    "None of above."\\
    \item SP-156\_CR: A driver is told to make a stop every half an hour for the engine to cool down, for five times. How long do the stops take?\\
    "Three hours."\\
    "Two and a half hours.",\\
    \textbf{"Two hours.",}\\
    "None of above."\\
\end{enumerate}

\end{document}